\documentclass[letterpaper, 10 pt, conference]{format/ieeeconf}  

\usepackage{amsmath}
\makeatletter
\def\maketag@@@#1{\hbox{\m@th\normalfont\normalsize#1}}
\makeatother
\usepackage{amsfonts}

\usepackage{amsthm}
\usepackage[ruled,vlined]{algorithm2e}
\usepackage{mathtools}
\usepackage{tabularx}
\usepackage{graphicx}
\usepackage{subfigure}
\usepackage{enumerate}
\usepackage{stfloats}
\usepackage{float}
\usepackage{url}
\usepackage{verbatim}
\usepackage[linkcolor=black,citecolor=black,urlcolor=black,colorlinks=true]{hyperref}
\usepackage{cite}
\usepackage{hyperref}
\usepackage{algorithmicx}
\usepackage{array}
\usepackage{multirow}
\usepackage{longtable}
\usepackage{rotating}
\usepackage{caption}
\usepackage{graphicx}

\DeclareCaptionLabelFormat{fig}{Fig. #2}
\graphicspath{{figures/}}
\IEEEoverridecommandlockouts
\let\oldtwocolumn\twocolumn
\renewcommand\twocolumn[1][]{%
	\oldtwocolumn[{#1}{
		\begin{center}
            \includegraphics[width=0.99\textwidth]{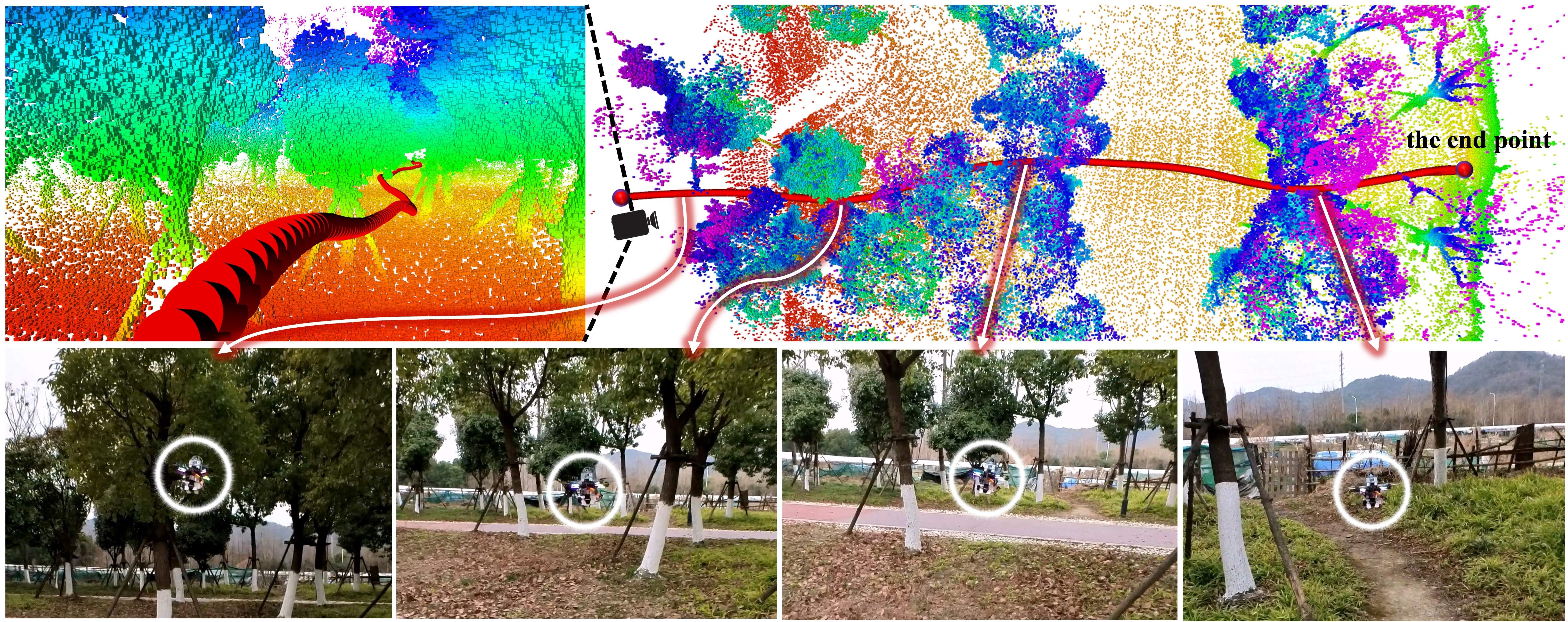}
			\captionsetup{font=footnotesize}
			\captionof{figure}{\textbf{Outdoor flight driven by an RL controller with onboard lidar sensing.}  The top part of the figure includes two different perspectives of the executed trajectory and point cloud maps generated by the estimation module in the system, Fast-LIO \cite{xu2021fast}. The left displays the flight trajectory (indicated by the red curve) from the view of its start point, while the right shows the flight trajectory from the top-down view. The bottom snapshots illustrate the drone's states at different progressions along the trajectory.}
			\label{fig:head}
		\end{center}
	}]
}

\begin{document}
	
	\author{Guangtong Xu$^\dag$, Tianyue Wu$^\dag$, Zihan Wang$^\dag$, Qianhao Wang, and Fei Gao
		\thanks{Guangtong Xu is with the Huzhou Institute, Zhejiang University, Huzhou 313000, China. (e-mail: guangtong\_xu@163.com).
			Tianyue Wu, Qianhao Wang, and Fei Gao are with the College of Control Science and Engineering, Zhejiang University, Hangzhou 310027, China, and the Huzhou Institute, Zhejiang University, Huzhou 313000, China. (e-mail: \{{tianyueh8erobot, qhwangaa, fgaoaa}\}@zju.edu.cn).
			Zihan Wang is with the Department of Automation, North China Electric Power University (Baoding), Baoding 071003, China, and the Huzhou Institute, Zhejiang University, Huzhou 313000, China. (e-mail: wangzh\_0111@163.com)}
		\thanks{$^\dag$Indicates equal contribution.}
	}
	
	\title{\LARGE \bf
		Flying on Point Clouds with Reinforcement Learning
	}

	\maketitle

	\begin{abstract}
		A long-cherished vision of drones is to autonomously traverse through clutter to reach every corner of the world using onboard sensing and computation. In this paper, we combine onboard 3D lidar sensing and sim-to-real reinforcement learning (RL) to enable autonomous flight in cluttered environments.  Compared to vision sensors, lidars appear to be more straightforward and accurate for geometric modeling of surroundings, which is one of the most important cues for successful obstacle avoidance.  On the other hand, sim-to-real RL approach facilitates the realization of low-latency control, without the hierarchy of trajectory generation and tracking. We demonstrate that, with design choices of practical significance, we can effectively combine the advantages of 3D lidar sensing and RL to control a quadrotor through a low-level control interface at 50Hz. The key to successfully learn the policy in a lightweight way lies in a specialized surrogate of the lidar's raw point clouds, which simplifies learning while retaining a fine-grained perception to detect narrow free space and thin obstacles. Simulation statistics demonstrate the advantages of the proposed system over alternatives, such as performing easier maneuvers and higher success rates at different speed constraints. With lightweight simulation techniques,  the policy trained in the simulator can control a physical quadrotor, where the system can dodge thin obstacles and safely traverse randomly distributed obstacles.
	\end{abstract}
	
\section{Introduction}
\label{sec:intro}
The past decade has witnessed the breakthroughs of autonomous drones from being protected by external localization devices \cite{mellinger2011minimum} to flying through clutter outside the laboratory with only onboard sensing and computing \cite{zhou2022swarm}. However, autonomous flight systems can still fail when faced with complex environments \cite{kong2021avoiding}, imperfect state estimation \cite{wu2024whole}, noisy environmental modeling, and high-latency decision making \cite{loquercio2021learning}.

With the rapid development of offboard computing and data generation \& utilization techniques, data-driven methods are expected to solve the above problems. For example, the expressive power of neural networks allows the burden of online inference to be reduced through offline training, thereby reducing control latency \cite{loquercio2021learning}. The need for separate state estimation and environment construction modules can also be reduced \cite{geles2024demonstrating, wu2024whole, zhang2024back}. One of the most user-friendly ways to obtain robust closed-loop policies is through sim-to-real \emph{interactive} learning methods \cite{ross2011reduction, kober2013reinforcement} that conduct training in simulation with massive data generation, such as sim-to-real reinforcement learning (RL) \cite{andrychowicz2020learning, hwangbo2019learning}. At the cost, these approaches fundamentally suffer from the discrepancies between simulation and the real world, which should be carefully handled to mitigate. For instance, employing state estimation \cite{kaufmann2020deep} or explicit mapping \cite{zhao2024learning} can alleviate the performance gap during deployment; by downsampling the high-dimensional exteroception \cite{miki2022learning,zhang2024back}, the controller can generalize well in specific domains and tasks.

Beyond the choice of methods for developing sensorimotor policy as mentioned above, the source of environmental perception is also a key design choice for autonomous flight. While most existing works use visual sensors, such as depth cameras, to model the environments, the sensory data produced by these sensors are typically noisy, which requires multi-frame fusion, e.g., through occupancy grid maps~\cite{zhou2022swarm}. Unfortunately, such approaches can mark grid cells containing thin obstacles as free regions due to massive non-obstacle observations around with a probabilistic-based formulation \cite{moravec1985high}, fundamentally resulting in the loss of detection of small objects.  In contrast, 3D lidars offer more comprehensive and accurate environmental perception. With high-resolution direct 3D perception and precise range measurements, lidars excel at directly detecting small obstacles~\cite{kong2021avoiding, ren2025safety}. Moreover, lidars are becoming smaller and more affordable, making them increasingly suitable for micro drone applications \cite{livoxmid360}.

In this paper, we combine 3D lidar sensing with the sim-to-real RL approach to implement a deployable autonomous flight system, which can smoothly fly through clutters, even with small obstacles. A neural controller mapping high-dimensional exteroception and proprioception to low-level control commands, e.g., collective thrusts and bodyrates, is trained using RL. However, the raw point clouds produced by lidars are enormous in volume, making RL from scratch challenging. Naive downsampling of point clouds to ease the burden of learning \cite{zhang2024back} risks losing detection of small objects~\cite{ren2025safety} or making it difficult to discern free areas amid dense obstacles~\cite{eldar1997farthest}. Therefore, the responses to fine-grained sensory information without excessive downsampling are particularly important when faced with dense or fine obstacles.  Motivated by this, we design a task-specific representation for point cloud aggregation as input to the policy, which is essentially a specialized downsampling of the raw point clouds and can divide the observed and unknown regions of surroundings to develop a policy with awareness of the unknown    \cite{zhao2024learning, tordesillas2021faster, ren2025safety}. Accurate lidar-based local state estimation, a lightweight yet effective perception simulation and dynamics domain randomization enable the policy to drive a micro quadrotor in the real world.

In summary, our main technical contributions are
\begin{itemize}
	\item design of a task-relevant lidar sensing representation that enables lightweight training of RL from scratch, while maintaining a high-resolution environmental perception;
	\item system integration of onboard lidar sensing and sim-to-real RL, and its demonstration of driving a physical quadrotor through clutter with fine obstacles.
\end{itemize}

\section{Related Work}
\label{sec:2A}
\subsection{Lidar-based Autonomous Flight in Clutter}
The problem of integrating perception data with motion planning has been extensively researched and practiced, manifested as the mapping module in the conventional drone navigation frameworks \cite{zhou2019robust,zhou2020ego,zhou2022swarm,tordesillas2021faster,han2021fast}. The most common approach for systems equipped with lidar sensors, similar to those using depth cameras, employs probabilistic occupancy grid maps \cite{moravec1985high,cai2023occupancy,ren2024rog} \footnote{In the article, we use both the terms of occupancy map or grid map to refer to methods based on the principles in \cite{moravec1985high}.}. However, this method, which traditionally conducts massive raycasting operations, incurs substantial computational costs when applied on lidar sensing that have extremely large spatial perception ranges. Moreover, this discretized spatial method may designate a grid cell containing thin obstacles as free when numerous other rays pass through the cell~\cite{ren2025safety}, thereby failing to effectively preserve lidar perception of thin obstacles.
Another approach is to directly perform motion planning on point cloud maps \cite{gao2019flying,zhang2019p,kong2021avoiding,wang2022geometrically,ren2022bubble,ren2025safety}, which maximally preserve the original lidar data information.
Compared to grid maps, since point clouds are inherently unordered and massive, direct extensive interaction with such maps is extremely time-consuming for onboard computation. Therefore, most point cloud map-based approaches extract simplified feasible spaces, also known as flight corridors \cite{ren2025safety,wang2022geometrically,ren2022bubble,wang2024fast}, to improve the efficiency for the motion planning algorithms' interaction with environmental perception. However, flight corridor implicitly filters environmental topological information, thereby losing the solution space. Notably, since point cloud maps cannot efficiently differentiate between known and unknown regions like grid maps, most point cloud map-based works consistently treat unknown areas as safe. To address this issue, the authors in \cite{ren2025safety} propose an convex extraction-based approach to distinguish known from unknown regions in point cloud maps using the calibrated FoV of the lidar sensor, and adopt the Faster strategy \cite{tordesillas2021faster} for the unknown region to enable safe navigation. In summary, traditional map representation methods for autonomous navigation appear suboptimal for high-precision, wide-range point cloud inputs from lidar sensors. Our proposed representation for lidar sensing can perserve the fine-grained sensory data and also naturally utilize the field-of-view (FoV) of the sensor to divide the observed and unknown region, without an explicit extraction of free space. \vspace{-0.1cm}

\subsection{Reinforcement Learning for Autonomous Flight in Clutter}
The method presented in \cite{sadeghi2016cad2rl} is the first efforts that use a sim-to-real RL method to control a drone to flight in clutter, where the raw RGB image is input and massive visual domain randomization is employed for successful sim-to-real transfer. However, the task scenario only includes hallways and the control interface is the vehicle's velocity rather than low-level control commands. Raw RGB input is also used in \cite{singla2019memory} for quadrotor obstacle avoidance, while the control commands are discretized into simple behaviors and the sim-to-real transfer is not demonstrated. The authors in \cite{song2023learning} use a teacher-student framework \cite{miki2022learning} to train a depth sensing-based policy to fly at high speed in a prior known environment. The authors in \cite{zhang2024back} and \cite{hu2024seeing} use differentiable simulators to provide first-order gradients for sample-efficient policy optimization to control a quadrotor using acceleration commands in clutter to follow a desired velocity without explicit position inputs, using depth sensing and monocular optical flow, respectively. However, they aggressively downsample the resolution of the images to 16 $\times$ 12 for generalizability, making the policy unaware of narrow feasible space or small obstacles. The authors in \cite{xu2024navrl} propose an RL velocity controller for RGB-D-sensing in dynamic environments with a velocity obstacle safety shield. Some works  \cite{kulkarni2024reinforcement,yu2024mavrl} use separate representation learning methods to enable RL from high-dimensional inputs to decide the acceleration commands for obstacle avoidance. Recently, there are works \cite{zhao2024learning,yu2024mavrl} proposing the idea to use RL for speed adaptation for flight in clutter (both of the mentioned works use a stereo depth sensor) according to the partially observed environment. The method in \cite{yu2024mavrl} obtains speed-adaptive policy by training the policy in varying complexity environments which incentivizes the behavior of speed adaptation. Distinguishing this work from those, the proposed system uses high-resolution 3D lidar sensing and  thrust and bodyrates for control of higher frequency (50Hz).

\begin{figure} \centering
	\subfigure {\includegraphics[width=0.48\textwidth]{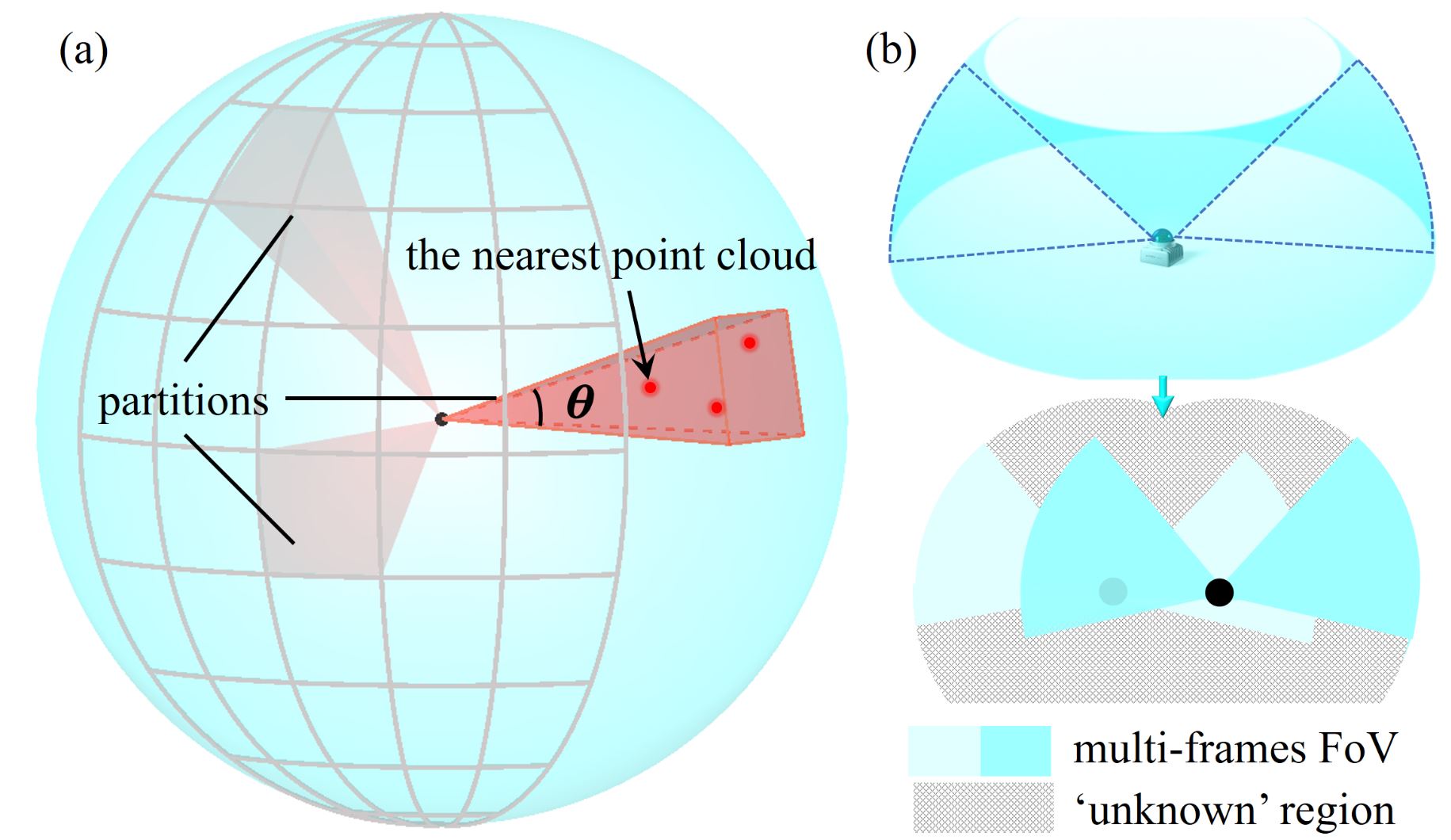}}
	\captionsetup{font=footnotesize}
	\caption{ \textbf{The illustration of the task-relevant lidar sensing representation.} (a)  The red square cones are examples of partitions for constructing the lidar sensing input, and the red dots represent the raw point cloud. The grid lines on the sphere visualize the partitions. (b) On the top is a schematic of the lidar's FoV, where the cartoon image is from the Livox Mid-360 website \cite{livoxmid360}. On the bottom is the unknown region calculated from the historical frames of FoV.}
	\label{fig:lidar_FOV}
	\vspace{-0.2cm}
\end{figure}

\begin{figure*} \centering
	\subfigure {\includegraphics[width=0.88\textwidth]{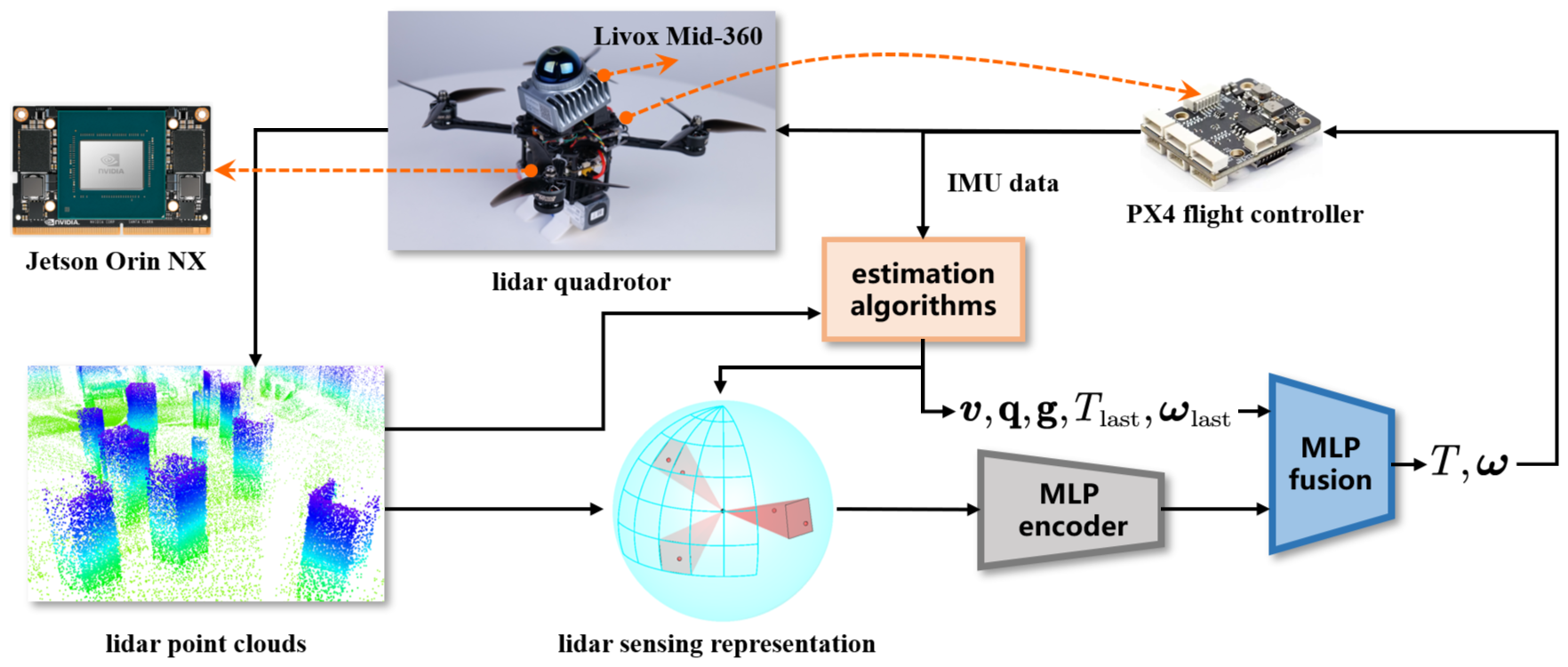}}
	\captionsetup{font=footnotesize}
	\caption{ \textbf{The system and policy architectures.} The system includes algorithmic components such as a MLP neural controller and estimation algorithms. The lidar sensing representation from MLP encoder, along with velocity $\boldsymbol{v}$, attitude $\boldsymbol{\mathrm q}$, goal direction $\boldsymbol{\mathrm g}$, last desired thrust $T_{\mathrm{last}}$, and last desired bodyrate $\boldsymbol \omega_{\mathrm{last}}$ are fed into the MLP fusion module. The output command is desired thrust $T$ and bodyrate $\boldsymbol \omega$.}
	\label{fig:network}
	\vspace{-0.2cm}
\end{figure*}

\section{Sim-to-Real RL for Flying on Point Clouds}
\label{sec:3}
We wish to control a quadrotor to safely fly through an environment with obstacles from one point to another with only lidar sensing and proprioception. In this section, we first describe how we implement a reinforcement learning pipeline, including a specialized observation space. We also detail the simulation setup and the sim-to-real techniques.

\subsection{Learning the Policy}
\label{sec:3A}
\subsubsection{Observation Space}
The observation space can be divided into an exteroception part from the lidar sensing and a proprioception part from an IMU and local state estimator \cite{xu2021fast}. We note that, in the real-world deployment, to ease the computation burden of constructing the exteroception representation, we first filter the raw point clouds far from the robot and uniformly downsample the point clouds at a resolution of 0.05m.

We illustrate the construction of the proposed representation in Fig. \ref{fig:lidar_FOV}. Specifically, the $k$ frames of history point clouds, where the point clouds are sampled at 10Hz, are transformed into the body frame at the current timestamp $t$ using the local state estimation, as shown in Fig. \ref{fig:network}. Then, centered on the robot, the space is divided into $n$ partitions at equal angles within its body frame, as illustrated in Fig. \ref{fig:lidar_FOV}(a). $n$ usually has a value of a few thousand. In our implementation, $n$ is taken to be 3200, which corresponds to an angular resolution of 4.5deg. Then, the distance (in meters) from the nearest point cloud, if at least one point is detected in the partition, to the robot is used as the \emph{value} of this partition, where the value over 10m is truncated to 10. For point cloud-free regions, we compute a $d_\text{unknown}$, which is the distance from the robot to the unknown region computed by the FoVs of historical $k$-frames lidar sensing as illustrated in Fig. \ref{fig:lidar_FOV}(b), where $0 < d_\text{unknown} < 10$. We set the value of this partition as $20-d_\text{unknown}$. The distinction of unknown regions inform the policy to perform a partial observability-aware behavior \cite{zhao2024learning}. The values of the partitions are listed as an $n$-dimension input to the policy.

The proprioception part includes the current ego-centric velocity given by the state estimator, attitude given by the IMU, the estimated height (z-axis position), and last taken action.  The normalized relative direction in the x-y plane towards the goal is also input the policy to inform the target.

We discuss the differences and advantages of our specialized surrogate of the point clouds over two alternative forms of point cloud inputs: (i) downsampled raw point clouds and (ii) occupancy map \cite{moravec1985high,miki2024learning,zhao2024learning}. A nice wish is to use the network to process high-dimensional raw sensor inputs and map them into low-level control commands. However, in our setup, the number of raw point clouds between a few frames reaches a magnitude of $10^5$ and we have to downsample them in a constant dimension for efficient RL training and alleviating the sim-to-real gap, typically by uniform sampling, random sampling, or farthest point sampling (FPS) \cite{eldar1997farthest}. However, when the environment harbors fine obstacles, it can be difficult for these task-agnostic sampling approaches to balance the burden of sample volume with the need to maintain detection of fine objects, and sample the point clouds into a constant volume. Moreover, the policy cannot tell the distinction between observed and unknown regions from the point cloud modality alone, which is demonstrated crucial for successful obstacle avoidance in some scenarios \cite{zhao2024learning,ren2025safety}. Another common RL input is occupancy maps transformed from point clouds or depth maps by local state estimation \cite{moravec1985high}. This approach, while mitigating the sim-to-real gap of sensor modeling compared to raw point cloud inputs and preserving the impact of fine obstacles, struggles with the trade-off between high dimensionality and accuracy. The problem of high dimensionality of this uniform space division approach is exacerbated by the long-range sensing of lidars. This is also detrimental to RL training, as with limited computation resource, we can only use a small mini-batch size for network updates, leading to sub-optimal performance \cite{berner2019dota}.  Our proposed method, on the other hand, essentially combines the ideas of downsampling and occupancy maps by dividing the perceived point clouds according to their \emph{importance}: the closer it is perceived, the more danger it may cause and the higher perception resolution it should be.  The perception of small obstacles is preserved due to a fine-grained partition, e.g., with a number of a few thousand. However, due to the fact that there is an information loss compared to occupancy maps, small obstacles in the distance may be perceived as larger ones. Empirically, this does not affect the safety of the system (see Sec. \ref{sec:5B}).

\subsubsection{Action Space}
The output of the policy is the desired thrust and bodyrates at 50Hz. These commands are executed by an embedded flight controller on the quadrotor.  This control interface allows for great maneuverability of the vehicle without introducing a large sim-to-real gap \cite{kaufmann2022benchmark,song2023reaching,wu2024whole}.

\subsubsection{Termination}
The agent only terminates when (i) its z-axis position falls outside a predefined range, which is set by the user according to the deployed scenario, and (ii) collision is detected between the vehicle and the environment, which is efficient using a pre-built inflated grid map. When the agent meets the termination condition, the quadrotor is reset to free space.

\subsubsection{Reward Function}
The reward function can be expressed as follows:
\begin{equation*}
	\begin{split}
		r=r_{\mathrm{forward}}+r_{\mathrm{thrust}}+r_{\mathrm{smoothness}}+r_{\mathrm{max} \_\mathrm{speed}}\\+r_{\mathrm{z}}+r_{\mathrm{ESDF}}+r_{\mathrm{collision}}+r_{\mathrm{yaw}},
	\end{split}
\end{equation*}
where $r_{\mathrm{forward}} =
\left\| \boldsymbol{p}_{\mathrm{goal}}-\boldsymbol{p} \right\| -\left\| \boldsymbol{p}_{\mathrm{goal}}-\boldsymbol{p}_{\mathrm{last}} \right\|$, $\boldsymbol{p}_{\mathrm{goal}}$ is the location of the navigation target,  $\boldsymbol{p_{\textnormal{last}}}$ is the position of the quadrotor in current timestamp, and $\boldsymbol{p}$ is the position of the quadrotor in the last timestamp, which encourages the quadrotor to fly towards the goal; $r_{\mathrm{thrust}}=\left\| T-g \right\|$, where $T$ is the magnitude of the collective thrust and $g$ is the magnitude of the gravity; $r_{\mathrm{smoothness}}=\left\| \boldsymbol{\omega} \right\| + \left\| \boldsymbol{a}-\boldsymbol{a}_{\mathrm{last}} \right\|$, where $\boldsymbol{\omega}=[\omega_x, \omega_y, \omega_z]$ and $\boldsymbol{a}=[\hat{T}, \hat{\omega_x}, \hat{\omega_y}, \hat{\omega_z}]$ are the bodyrate and the output of the policy at the current timestamp, respectively, and $\boldsymbol{a}_{\mathrm{last}}$ is the last taken action, encouraging a smooth locomotion; $r_{\mathrm{max} \_\mathrm{speed}} = -e^{\max\{0, \left\| \boldsymbol{v} \right\|-v_{\mathrm{max}}\}} + 1$, where $\boldsymbol{v}=[v_x, v_y, v_z]$ is the velocity of the quadrotor and $v_{\mathrm{max}}$ is the user-defined velocity constraint; $r_z = \max\{z-z_{\mathrm{max}}, z_{\mathrm{min}}-z, 0\}$, where $z$ is the z coordination of the quadrotor in the world frame; $r_{ESDF}=\lambda(1 - e^{-k d^2})$, where $d$ is the distance from the robot to the nearest point in the environment, and $\lambda>0$ and $k>0$ are some parameters, which is a shaping reward for the sparse collision penalty to encourage the quadrotor to move far from the obstacles; $r_{\mathrm{collision}}=-10$ applied only when collision is detected; $r_{\mathrm{yaw}}=\boldsymbol{x}_{\mathrm{body}} \cdot \boldsymbol{v}/\left\|\boldsymbol{v}\right\|$, where $\boldsymbol{x}_{\mathrm{body}}$ is the normalized vector of x-axis of the body frame and $\cdot$ denotes dot product, which is to encourage the head of the quadrotor to align with its movement direction. We note that the weight of each reward component is omitted  in the above formulation.

\subsubsection{Policy Representation}
The policy consists of a multilayer perceptron (MLP) encoder to encode the 3200-dimensional exteroceptive input into a 128-dimension hidden state and an MLP fusion module to fuse the hidden state and other vectorized inputs. The MLP encoder is implemented with 3 layers, where the output dimensions of the hidden states are 128, 64, 64, respectively. The fusion module is implemented with 4-layers MLPs, where the output dimensions of the hidden states are 128, 256, 256, 128, respectively. Then a projection layer maps the hidden state into the 4-dimensional action. The neural network architecture is illustrated in Fig. \ref{fig:network}.

\subsection{Simulation}
\label{sec:3a}
For quadrotor dynamics, we use the air drag augmented model used in the open-source codes of \cite{heeg2024learning}. We also simulate the motor delay as a first-order system.
\subsubsection{Dynamics Domain Randomization}
The actuation of micro drones is noisy and complex aerodynamics effects make it almost impossible to simulate the perfect dynamics transition in a lightweight way. Therefore, we apply specialized dynamics domain randomization for developing a robust policy. Before being input to the model, the thrust and bodyrates are randomized by $\pm$10$\%$ and $\pm$8$\%$ \cite{kaufmann2022benchmark}. We note that the randomization coefficients around 5$\%$ to 10$\%$ should work similarly. Instead of identifying the drag coefficients accurately, we largely randomize them at $\pm$30$\%$, since we have the explicit state estimation of velocity to ensure a reasonable performance of the policy in simulation. We find that such an approach of brute force helps the exploration of the RL agent during training.

\subsubsection{Lidar Sensing Simulation}
Inspired by an effective lightweight lidar simulator~\cite{kong2023marsim}, we extract the unique spatio-temporal patterns of a specific lidar device, by fitting a parameterized representation of the real-world lidar's polar coordinates in the case where we sample the lidar data at 10Hz, instead of directly applying the simulated point clouds in the calibrated FoV. Then, in our own simulation environment, we apply these polar coordinates to directly index corresponding points on the pre-built obstacle surfaces represented by point clouds with a resolution of 0.05m, thereby simulating the lidar sensor data.

\begin{figure} \centering
	\subfigure {\includegraphics[width=0.46\textwidth]{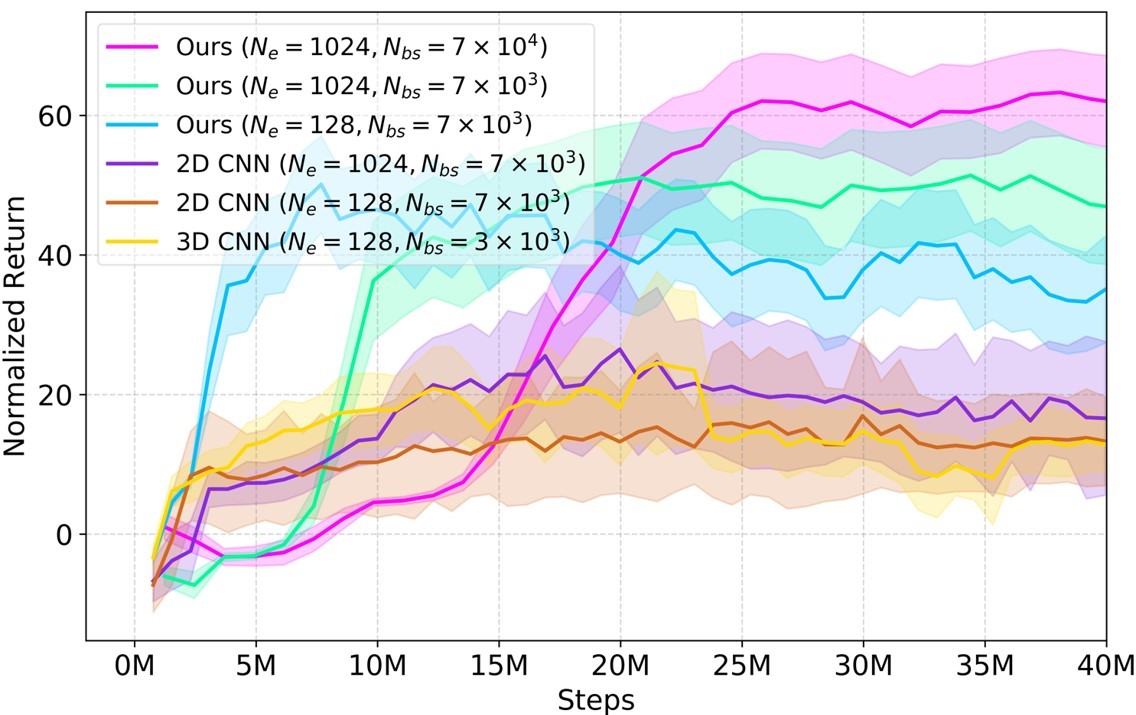}}
	\captionsetup{skip=5pt}
	\captionsetup{font=footnotesize}
	\caption{\textbf{The evolution of undiscounted return of the proposed exteroception representation vs occupancy map during training.} $N_e$ and $N_{bs}$ denote the number of training environments and mini-batch size, respectively.}
	\label{fig:alat}
	\vspace{-0.3cm}
\end{figure}

\section{System and Implementation}
\subsection{System Overview \& Setup}
The overview of the integrated system is in Fig. \ref{fig:network}. Specifically, a Livox Mid-360 lidar sensor is equipped on the quadrotor for onboard sensing. The outputs of the policy are executed by an onboard PX4 Autopilot flight controller\footnote{https://px4.io/}. The data from the  IMU on the flight controller and the point clouds are fused according to \cite{xu2021fast} to generate high-frequency state estimation that is locally very accurate.  The  flight controller is elaborately tuned to respond to the commands with bodyrates' delay less than 20ms. We use a Jetson Orin NX\footnote{https://www.nvidia.com/en-us/autonomous-machines/embedded-systems/jetson-orin/}  as the onboard computer for the computation of estimation algorithms and GPU inference.

\subsection{Training Implementation}
\label{sec:4b}
The policy is trained with the  proximal policy optimization (PPO) algorithm \cite{schulman2017proximal}. We use 1024 environments to collect data of 300 time steps between two policy updates. Thanks to the relatively low-dimensionality of the designed representations for exteroception inputs, we can use a much larger batch size of 70000 relative to that available when using occupancy maps (see Sec. \ref{sec:5A}) with the same computation device.  The weights of actor and critic networks are not shared. Data collection is conducted on an i9-14900K CPU and the neural network optimization is on an NVIDIA RTK 4090 GPU.

\begin{figure*} \centering
	\subfigure {\includegraphics[width=1.0\textwidth]{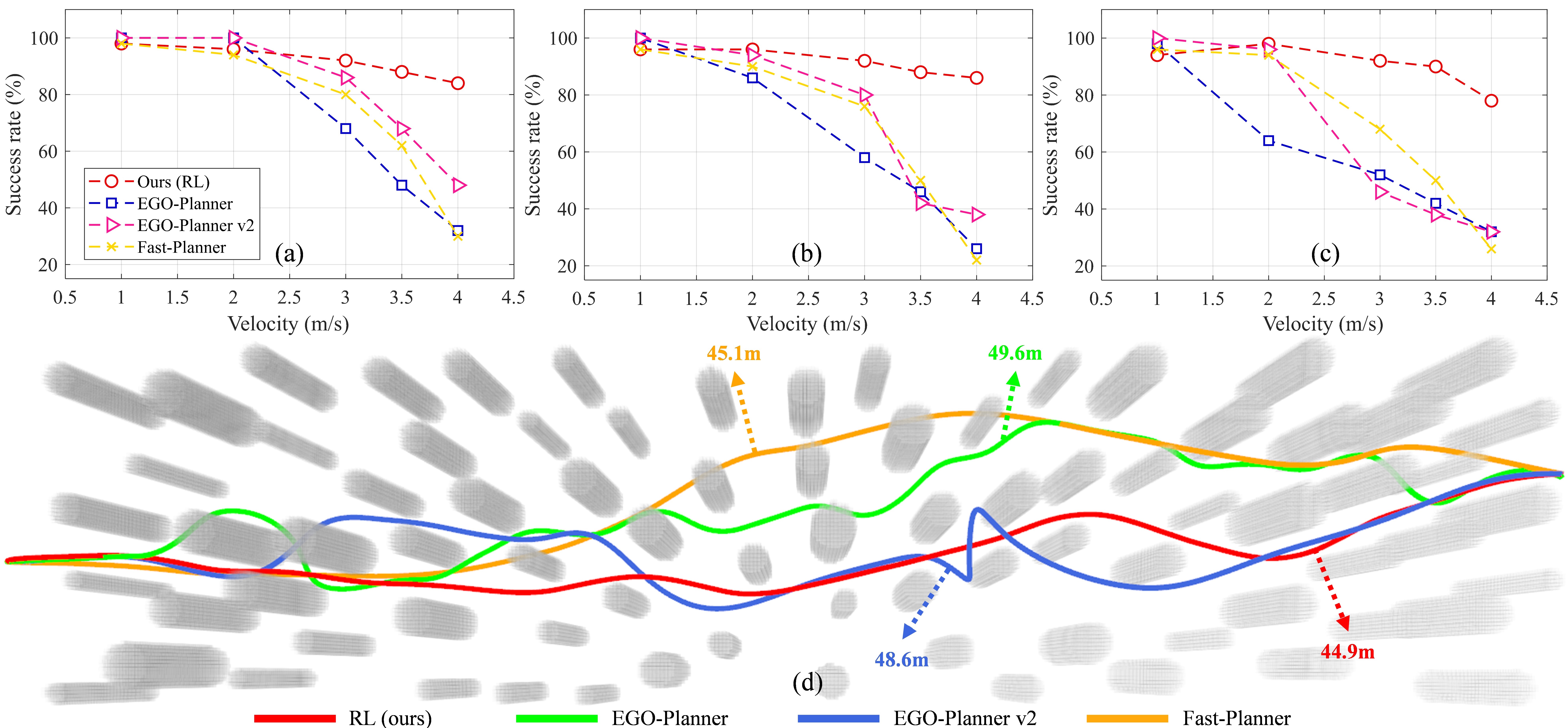}}
	\captionsetup{skip=5pt}
	\captionsetup{font=footnotesize}
	\caption{\textbf{Benchmark results with previous systems under different maximum speeds.} Figures (a)-(c) illustrate the comparison results of success rate in scenarios I, II, and III, respectively. Scenario I: A 40m$\times$10m map with around 100 obstacles, and the obstacle radius is randomly sampled within 0.5 $\sim$ 0.7m.
		Scenario II: The size of the obstacles is set within 0.5 $\sim$ 0.7m, but the number of obstacles is increased to around 130.
		Scenario III: The obstacle radius is randomized between 0.5m $\sim$ 1.2m, with around 90 obstacles. (d) The example (successful) trajectory results of the compared approaches in scenario III with the speed constraint of 3.0m/s, and the corresponding flight distances are provided.}
	\label{fig:benchmark}
	\vspace{-0.2cm}
\end{figure*}

\section{Results}
In this section, we conduct experiments in both simulation and the real world, and demonstrate that
\begin{itemize}
	\item the designed representation as input enables lightweight learning, while using occupancy maps struggles to effectively learn the policy  from scratch (Sec. \ref{sec:5A});
	\item massive data generation with large mini-batch size (in our limited computation resource) crucially contribute to our method (Fig. \ref{fig:alat});
	\item the proposed method demonstrates superior performance in simulation when benchmarked with widely used open-source systems, quantified as success rates at different maximum speeds (Sec. \ref{sec:5B});
	\item the success in simulation can be reproduced in the real world (Sec. \ref{sec:5C}).
\end{itemize}

\subsection{The Proposed Exteroception Representation vs Occupancy Map}
\label{sec:5A}
\subsubsection{Setup} In this subsection, we employ the occupancy map for comparison with our proposed representation. The occupancy map can also reveal thin obstacles and unknown areas as described in Sec. \ref{sec:3A}. The resolution in each axis is set as 0.2m. We use 3-layers 3D convolutional neural networks (CNNs) and 2D CNNs (both with 2 layers of MLPs) for processing 3D occupancy maps into 128-dimension features, respectively. The 2D CNN encoder treats the grid units along the z-axis as the channel dimension.  We run training and evaluation on a fixed environment distribution for a controllable testbed to track the evolution of the performance during training. The results are illustrated in Fig. \ref{fig:alat}. We employ various parameter settings about the number of environments for collecting data, denoted as $N_\textnormal{e}$ in Fig. \ref{fig:alat}, and the mini-batch size, denoted as $N_\textnormal{bs}$. The speed constraint is set as 3.0m/s and the historical frame number $k$ is set as 5. The training setting is consistent with that described in Sec. \ref{sec:4b}.

\subsubsection{Results}The results show, a bit surprisingly, the naive application of the PPO algorithm on occupancy maps makes it difficult to establish an effective policy from scratch. We hypothesize that this result is caused by the high-dimensionality of the input and the end-to-end training scheme. A supervised representation learning method \cite{stooke2021decoupling, kulkarni2024reinforcement, yu2024mavrl} may be able to alleviate such a problem by decoupling the two issues of RL, i.e., exploration and high-dimensional observation-to-action mappings, which is beyond the scope of this work. Moreover, at a resolution of 0.2m, the dimension of the occupancy map is 50$\times$50$\times$15, which makes the maximum mini-batch size that can be used about the magnitude of $1.5\times 10^4$ for 2D CNN and $5\times 10^3$ for 3D CNN in our implementation on a GPU with a 24G memory. Such a relatively small mini-batch size can degrade the performance of PPO, as observed in \cite{berner2019dota}.   On the contrary, the employed exteroception representation, although still at thousands of dimensions, enables a larger mini-batch size over the magnitude of $1\times 10^5$ and a successful implementation of end-to-end RL training without excessive tuning.

\begin{figure} \centering
	\subfigure {\includegraphics[angle=270,width=\linewidth]{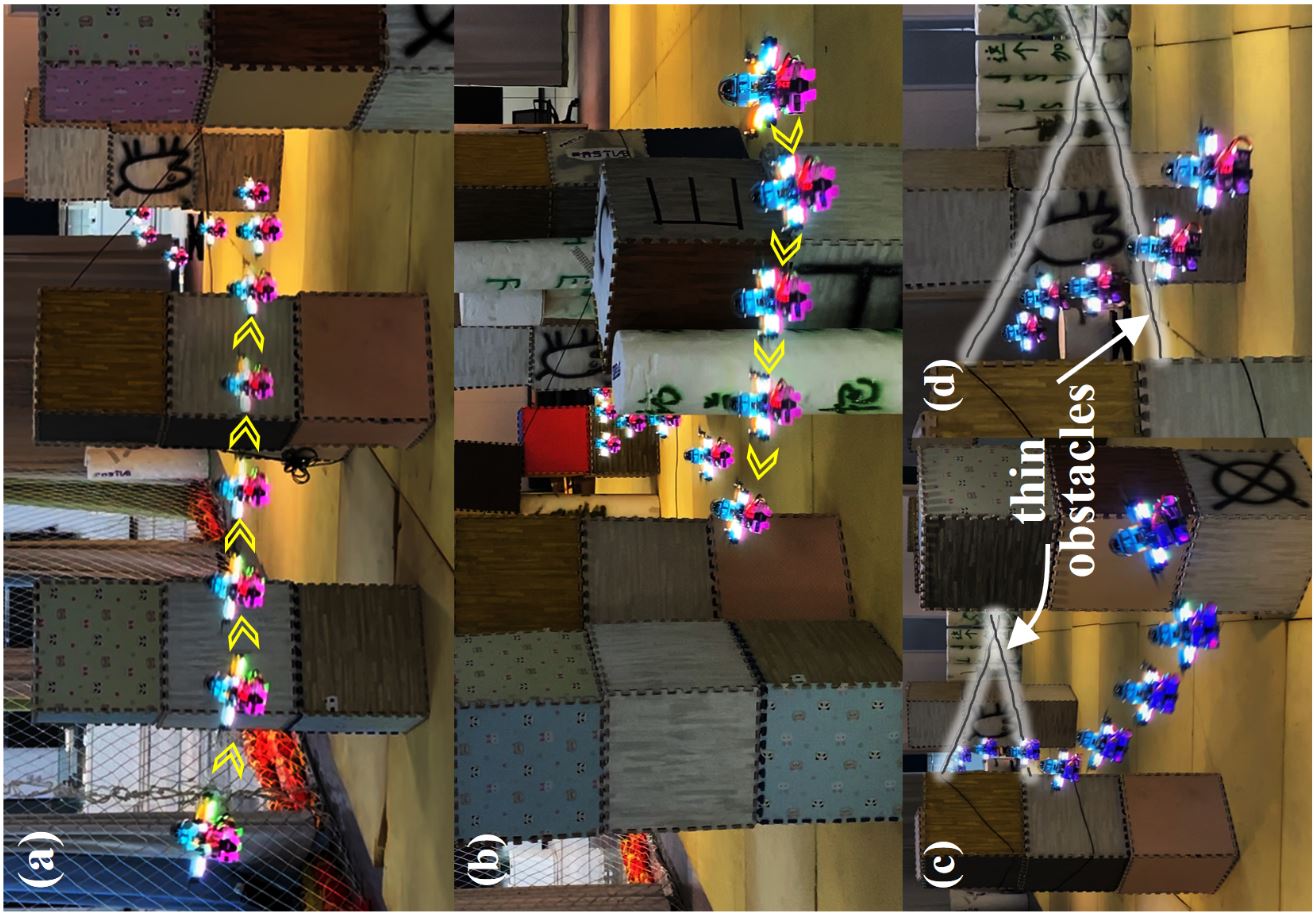}}
	\captionsetup{skip=5pt}
	\captionsetup{font=footnotesize}
	\caption{\textbf{Results of indoor flight of the proposed system.} (a) Indoor scenario I. (b) Indoor scenario II. The quadrotor can fly through the cluttered environments safely. The arrows represent the flight direction. (c) \& (d) Two highlighted trials of thin-obstacle avoidance and corresponding snapshots.}
	\label{fig:indoor}
	\vspace{-0.3cm}
\end{figure}

\subsection{Benchmark with Previous Systems in Simulation}
\label{sec:5B}
\subsubsection{Setup}We benchmark our system with three representative previous obstacle avoidance systems: Fast-Planner \cite{zhou2019robust}, EGO-Planner \cite{zhou2020ego}, and EGO-Planner v2\cite{zhou2022swarm}. Fast-Planner and EGO-Planner are among the most widely used open-sourced navigation systems in unknown environments. Fast-Planner applies an unconstrained trajectory optimization on an online updated Euclidean Signed Distance Field (ESDF). EGO-Planner is developed based on a local and greedy trajectory optimization method directly applied to the occupancy map without establishing ESDF with an elaborately designed finite state machine (FSM). The ESDF-free design of EGO-Planner reduces decision-making latency compared to alternative systems \cite{zhou2019robust}. Ego-Planner v2 employs the MINCO trajectory representation \cite{wang2022geometrically} instead of the B-spline used in vanilla EGO-Planner, which improves the quality and efficiency of trajectory generation.

For a fair comparison, the sensing FoV and frequency of the benchmarked methods are set the same as those in this work. The statistics results are illustrated in Fig. \ref{fig:benchmark}. The results are calculated under various speed constraints with 50 trials. The test scenarios are generated randomly according to specific parameterized settings: the obstacle radius in test scenario I is randomly sampled from 0.5m to 0.7m, distributed in a 40m$\times$10m map with the number of obstacles around 100; the size of obstacles in test scenario II is consistent with the first scenario, while the number of obstacles around 130; the obstacle radius in scenario III is largely randomized among 0.5m to 1.2m, while the number of obstacles is around 90. The policy is not trained on the evaluation environments. The $k$ is set as 5 in our experiments.

\subsubsection{Results}
In low-speed domains, e.g., the maximum speed  is set as 1m/s, all compared methods perform well. However, when the speed limitations are relaxed,  the performance of them degrades drastically, due to their greedy trajectory optimization design and handcrafted FSMs, especially for EGO-Planner. The online optimization also causes a significant perception latency with limited computational sources \cite{loquercio2021learning}. The sub-optimal designs in these systems result in the occasional sacrifice of dynamical feasibility by the trajectory optimization methods \cite{zhao2024learning} and the compounding error of imperfect trajectory tracking control \cite{song2023reaching,wu2024whole}, which also fundamentally contribute to the eventual failure.  A similar phenomenon is observed in \cite{zhao2024learning}.  At the same time, the benchmarked systems lack awareness of the partially observable nature of the task and rely only on high-frequency replanning to mitigate the impact caused by unknown environments. In high-speed settings, such an effect can be largely amplified under a constant perception frequency. RL's formulation alleviates the above problems and mitigates overly greedy optimization by massively sampling. From Fig. \ref{fig:benchmark}(c), the success rate by ours method maintains around 80\%,  while other methods are below 40\% when the speed constraint is 4.0m/s. Moreover, from Fig. \ref{fig:benchmark}(d), we can see that our policy executes an easier trajectory and traverses the environment along a shorter path compared to the EGO-Planner family. \vspace{-0.1cm}

\subsection{Real-world Demonstrations}
\label{sec:5C}
In this subsection, we conduct real-world experiments in both artificial clutters with fine obstacles and outdoor natural clutters.

\subsubsection{Safe Traversal through Boxes and Wires}
Inspired by recent impressive model-based lidar navigation works~\cite{kong2021avoiding,ren2025safety}, we design indoor tests with fine obstacles, i.e., wires with a diameter of 10 mm, where a vision-based commercial drone fails to avoid these obstacles \cite{ren2025safety}. Fig. \ref{fig:indoor} shows the trajectory rollouts by the policy in the scenario. The maximum speed constraint of the policy is set to 2.0m/s.

In particular, in both Fig. \ref{fig:indoor} (c) and (d), we observe an upward dodge over the wire obstacles. Thanks to the long-range FoV of the lidar sensing, the quadrotor is able to demonstrate an easy maneuver, smoothly flying through the free areas between the box-shaped obstacles.

\subsubsection{Safe Traversal through Trees}
We deploy the system in an outdoor scenario with unevenly distributed trees, as shown in Fig. \ref{fig:head}. Compared to indoor experiments, outdoor experiments are challenging not only due to the broad scene, where parts of the environment frequently exceed the sensing range resulting in massive invalid measurements, but also due to the unmodeled wind perturbations. We hypothesize the policy can handle these unmodeled factors via a large dynamics domain randomization, as described in \ref{sec:3a}. We train a policy with the maximum speed constraint as 3.0m/s. In the experimented scenario, the quadrotor flies over 25m in different trials. \vspace{-0.1cm}

\section{Conclusion, Limitations and Future Work}
In this work, we present one of the first efforts to combine 3D lidar sensing and sim-to-real RL for autonomous drone flights. We demonstrate that, without excessive downsampling, we are able to retain the perception of small obstacles and narrow spaces while achieving lightweight access to deployable policies without a hierarchy of trajectory generation and tracking. Thanks to the sound simulation techniques and accurate lidar-based state estimation, the policy can be deployed in the real world.

However, we find that the quadrotor cannot always avoid thin obstacles in the real world. First, this may be because the reflectivity varies among different materials, which may result in a loss of detection by the lidar's rays in some cases. Algorithmically, we suspect this is attributed to the idealized simulation methods currently used for lidar sensing, which directly extract point clouds from pre-generated environments based on the sensor's pattern. This introduces an unexpected sim-to-real gap compared to the point clouds generated by specific lidar sensors (e.g., the Livox Mid-360) in the real world. Specifically, we observed that even when facing a wall that theoretically should be perfectly observable by the lidar, the data obtained from the sensor contains numerous radius values of 0 in polar coordinates, indicating many invalid observations. These invalid measurements, however, are not simulated in our current simulation environment. At the same time, we find that in scenarios with particularly large-sized obstacles, the algorithm still tends to greedily acquire the rewards by a fastest-forward locomotion under the speed constraint and cannot be trained with a uniform parameter setup to produce a policy that can succeed in all the simulated scenarios.

In the future, we plan to implement an effective yet efficient simulation method for lidar sensing, which is still an open problem \cite{kong2023marsim}, and find out an improved training recipe or policy architecture to support a generalizable policy in the open world.

\newlength{\bibitemsep}\setlength{\bibitemsep}{0.00\baselineskip}
\newlength{\bibparskip}\setlength{\bibparskip}{0pt}
\let\oldthebibliography\thebibliography
\renewcommand\thebibliography[1]{
	\oldthebibliography{#1}
	\setlength{\parskip}{\bibitemsep}
	\setlength{\itemsep}{\bibparskip}
}
\bibliography{references}

\end{document}